\documentclass[runningheads]{llncs}
\usepackage{graphicx}
\usepackage{comment}
\usepackage{amsmath,amssymb} 
\usepackage{color}
\usepackage{multirow}

\begin{document}
\pagestyle{headings}
\mainmatter
\def\ECCVSubNumber{1018}  

\title{Decoupled Spatial-Temporal Attention Network for Skeleton-Based Action Recognition} 

\titlerunning{DSTA-Net for Skeleton-Based Action Recognition}
%
\author{Lei Shi$^{1,2}$, Yifan Zhang$^{1,2}$, Jian Cheng$^{1,2,3}$ and Hanqing Lu$^{1,2}$}
\authorrunning{Lei Shi et al.}
%
\institute{$^{1}$NLPR \& AIRIA, Institute of Automation, Chinese Academy of Sciences\\ 
$^{2}$School of Artificial Intelligence, University of Chinese Academy of Sciences, Beijing 100049, China \\ 
$^{3}$CAS Center for Excellence in Brain Science and Intelligence Technology\\
}
\maketitle

\begin{abstract}
Dynamic skeletal data, represented as the 2D/3D coordinates of human joints, has been widely studied for human action recognition due to its high-level semantic information and environmental robustness. 
However, previous methods heavily rely on designing hand-crafted traversal rules or graph topologies to draw dependencies between the joints, which are limited in performance and generalizability. 
In this work, we present a novel decoupled spatial-temporal attention network (DSTA-Net) for skeleton-based action recognition.  
It involves solely the attention blocks, allowing for modeling spatial-temporal dependencies between joints without the requirement of knowing their positions or mutual connections. 
Specifically, to meet the specific requirements of the skeletal data, 
three techniques are proposed for building attention blocks,
namely, spatial-temporal attention decoupling, decoupled position encoding and spatial global regularization. 
Besides, from the data aspect, we introduce a skeletal data decoupling technique to emphasize the specific characteristics of space/time and different motion scales, 
resulting in a more comprehensive understanding of the human actions.
To test the effectiveness of the proposed method, extensive experiments are conducted on four challenging datasets for skeleton-based gesture and action recognition, namely, SHREC, DHG, NTU-60 and NTU-120, where DSTA-Net achieves state-of-the-art performance on all of them. 

\end{abstract}   

\section{Introduction}
\label{sec:introduction}
Human action recognition has been studied for decades since it can be widely used for many applications such as human-computer interaction and abnormal behavior monitoring~\cite{carreira_quo_2017,shi_gesture_2019,feichtenhofer_slowfast_2019,shi_action_2019}. 
Recently, skeletal data draws increasingly more attention because it contains higher-level semantic information in a small amount of data and has strong adaptability to the dynamic circumstance~\cite{yan_spatial_2018,shi_two-stream_2019,shi_skeleton-based_2019}. 

The raw skeletal data is a sequence of frames each contains a set of points.
Each point represents a joint of human body in the form of 2D/3D coordinates. 
Previous data-driven methods for skeleton-based action recognition rely on manual designs of traversal rules or graph topologies to transform the raw skeletal data into a meaningful form such as a point-sequence, a pseudo-image or a graph, so that they can be fed into the deep networks such as RNNs, CNNs and GCNs for feature extraction~\cite{yan_spatial_2018,qiu_learning_2017,zhang_view_2017}. 
However, there is no guarantee that the hand-crafted rule is the optimal choice of modeling global dependencies of joints, 
which limits the performance and generalizability of previous approaches. 
Recently, transformer~\cite{vaswani_attention_2017,dai_transformer-xl:_2019} has achieved big success in the NLP field, whose basic block is the self-attention mechanism. 
It can learn the global dependencies between the input elements  with less computational complexity and better parallelizability. 
For skeletal data, employing the self-attention mechanism has an additional advantage that there is no requirement of knowing a intrinsic relations between the elements, thus it provides more flexibility for discovering useful patterns. 
Besides, since the number of joints of the human body is limited, the extra cost of applying self-attention mechanism is also relatively small. 

Inspired by above observations, we propose a novel decoupled spatial-temporal attention networks (DSTA-Net) for skeleton-based action recognition. 
It is based solely on the self-attention mechanism, without using the structure-relevant RNNs, CNNs or GCNs. 
However, it is not straightforward to apply a pure attention network for skeletal data as shown in following three aspects: 
(1) The input of original self-attention mechanism is the sequential data, while the skeletal data exists in both the spatial and temporal dimensions. 
A naive method is simply flattening the spatial-temporal data into a single sequence like~\cite{wang_non-local_2018}. 
However, it is not reasonable to treat the time and space equivalently because they contain totally different semantics~\cite{feichtenhofer_slowfast_2019}. 
Besides, simple flattening operation increases the sequence length, which greatly increases the computation cost due to the dot-product operation of the self-attention mechanism. 
Instead, we propose to decouple the self-attention mechanism into the spatial attention and the temporal attention sequentially.
Three strategies are specially designed to balance the independence and the interaction between the space and the time. 
(2) There are no predefined orders or structures when feeding the skeletal joints into the attention networks. 
To provide unique markers for every joint, a position encoding technique is introduced. 
For the same reason as before, it is also decoupled into the spatial encoding and the temporal encoding.
(3) 
It has been verified that adding proper regularization based on prior knowledge can effectively reduce the over-fitting problem and improve the model generalizability. 
For example, due to the translation-invariant structure of images, CNNs exploit the local-weight-sharing mechanism to force the model to learn more general filters for different regions of images. 
As for skeletal data, each joint of the skeletons has specific physical/semantic meaning (e.g., head or hand), which is fixed for all the frames and is consistent for all the data samples. 
Based on this prior knowledge, a spatial global regularization is proposed to force the model to learn more general attentions for different samples. 
Note the regularization is not suitable for temporal dimension because there is no such semantic alignment property. 

Besides, from the data aspect, the most discriminative pattern is distinct for different actions. 
We claim that two properties should be considered. 
One property is whether the action is motion relevant or motion irrelevant, which aims to choose the specific characters of space and time. 
For example, to classify the gestures of ``waving up'' versus ``waving down'', the global trajectories of hand is more important than hand shape, but when recognizing the gestures like ``point with one finger'' versus ``point with two finger'', the spatial pattern is more important than hand motion. 
Based on this observation, we propose to decouple the data into the spatial and temporal dimensions, where the spatial stream contains only the motion-irrelevant features and temporal stream contains only the motion-relevant features. 
By modeling these two streams separately, the model can better focus on spatial/temporal features and identity specific patterns. 
Finally, by fusing the two streams, it can obtain a more comprehensive understanding of the human actions. 
Another property is the sensibility of the motion scales. 
For temporal stream, the classification of some actions may rely on the motion mode of a few consecutive frames while others may rely on the overall movement trend. 
For example, to classify the gestures of ``clapping'' versus ``put two hands together'', the short-term motion detail is essential. 
But for ``waving up'' versus ``waving down'', the long-term motion trend is more important. 
Thus, inspired by \cite{feichtenhofer_slowfast_2019}, we split the temporal information into a fast stream and a slow stream based on the sampling rate. 
The low-frame-rate stream can capture more about global motion information and the high-frame-rate stream can focus more on the detailed movements. 
Similarly, the two streams are fused to improve the recognition performance. 

We conduct extensive experiments on four datasets, including two hand gesture recognition datasets, i.e., SHREC and DHG, and two human action recognition datasets, i.e., NTU-60 and NTU-120.
Without the need of hand-crafted traversal rules or graph topologies, our method achieves state-of-the-art performance on all these datasets, which demonstrates the effectiveness and generalizability of the proposed method. 

Overall, our contributions lie in four aspects: 
\begin{enumerate}
    \item To the best of our knowledge, we are the first to propose a decoupled spatial-temporal attention networks (DSTA-Net) for skeleton-based action recognition, which is built with pure attention modules without manual designs of traversal rules or graph topologies. 
    \item We propose three effective techniques in building attention networks to meet the specific requirements for skeletal data, namely, spatial-temporal attention decoupling, decoupled position encoding and spatial global regularization.  
    \item We propose to decouple the data into four streams, namely, spatial-temporal stream, spatial stream, slow-temporal stream and fast-temporal stream, each focuses on a specific aspect of the skeleton sequence. By fusing different types of features, the model can have a more comprehensive understanding for human actions. 
    \item On four challenging datasets for action recognition, our method achieves state-of-the-art performance with a significant margin. 
    DSTA-Net outperforms SOTA 2.6\%/3.2\% and 1.9\%/2.9\% on 14-class/28-class benchmarks of SHREC and DHG, respectively. 
    It achieves 91.5\%/96.4\% and 86.6\%/89.0\% on CS/CV benchmarks of NTU-60 and NTU-120, respectively. 
    The code will be released for future works. 
\end{enumerate}

\section{Related Work}

\textbf{Skeleton-based action recognition} has been widely studied for decades. The main-stream methods lie in three branches: 
(1) the RNN-based methods that formulate the skeletal data as a sequential data with a predefined traversal rules, and feed it into the RNN-based models such as the  LSTM~\cite{zhang_view_2017,li_independently_2018,si_skeleton-based_2018,si_attention_2019}; 
(2) the CNN-based methods that convert the input skeletons into a pseudo-image with hand-crafted transformation rules, and model it with various successful networks used in image classification fields~\cite{li_skeleton-based_2017,liu_enhanced_2017,cao_skeleton-based_2018};
(3) the GCN-based methods that encode the skeletal data into a predefined spatial-temporal graph, and model it with the graph convolutional networks~\cite{yan_spatial_2018,tang_deep_2018,shi_two-stream_2019}.
In this work, instead of formulating the skeletal data into the images or graphs, we directly model the dependencies of joints with pure attention blocks. 
Our model is more concise and general, without the need of designing hand-crafted transformation rules, and it outperforms the previous methods with a significant margin. 

\textbf{Self-attention mechanism} is the basic block of transformer~\cite{vaswani_attention_2017,dai_transformer-xl:_2019}, which is the mainstream method in the NLP field. 
Its input consists of a set of queries $Q$, keys $K$ of dimension $C$ and values $V$, which are packaged in the matrix form for fast computation. 
It first computes the dot products of the query with all keys, divides each by $\sqrt{C}$, and applies a softmax function to obtain the weights on the values~\cite{vaswani_attention_2017}. 
In formulation: 
\begin{equation}
    Attention(Q, K, V) = softmax(\frac{QK^T}{\sqrt{C}})
\end{equation}
The similar idea has also been used for many computer vision tasks such as relation modeling~\cite{santoro_simple_2017}, detection~\cite{hu_relation_2018} and semantic segmentation~\cite{fu_dual_2019}. 
To the best of our knowledge, we are the fist to apply the pure attention networks for skeletal data and further propose several improvements to meet the specific requirements of skeletons.

\section{Methods}
In this section, we first propose three techniques for building attention blocks in Sec.~\ref{sec:strategies}, Sec.~\ref{sec:positionencoding} and Sec.~\ref{sec:regularization}. 
Then, the basic attention module and the detailed architecture of the network are introduced in Sec.~\ref{sec:attention} and Sec.~\ref{sec:architecture}. 
Finally, the data decoupling and the overall multi-stream solution is described in Sec.~\ref{sec:decouple}. 


\subsection{Spatial-temporal attention module}
\label{sec:strategies}
Original transformer is fed with the sequential data, i.e., a matrix $X\in\mathbb{R}^{N\times C}$, where N denotes the number of elements and C denotes the number of channels. 
For dynamic skeletal data, the input is a 3-order tensor $X\in\mathbb{R}^{N\times T\times C}$, where T denotes the number of frames. 
It is worth to investigate how to deal with the relationship between the time and the space. 
Wang et al.~\cite{wang_non-local_2018} propose to ignore the difference between time and space, and regard the inputs as a sequential data $X\in\mathbb{R}^{\hat{N}\times C}$, where $\hat{N}=NT$. 
However, the temporal dimension and the spatial dimension are totally different as introduced in Sec.~\ref{sec:introduction}.
It is not reasonable to treat them equivalently. 
Besides, the computational complexity of calculating the attention map in this strategy is $O(T^2N^2C)$ (using the naive matrix multiplication algorithm), which is too large. 
Instead, we propose to decouple the spatial and temporal dimensions, where the computational complexity is largely reduced and the performance is improved. 

\begin{figure}[tp]
    \centering
    \includegraphics[width=\linewidth]{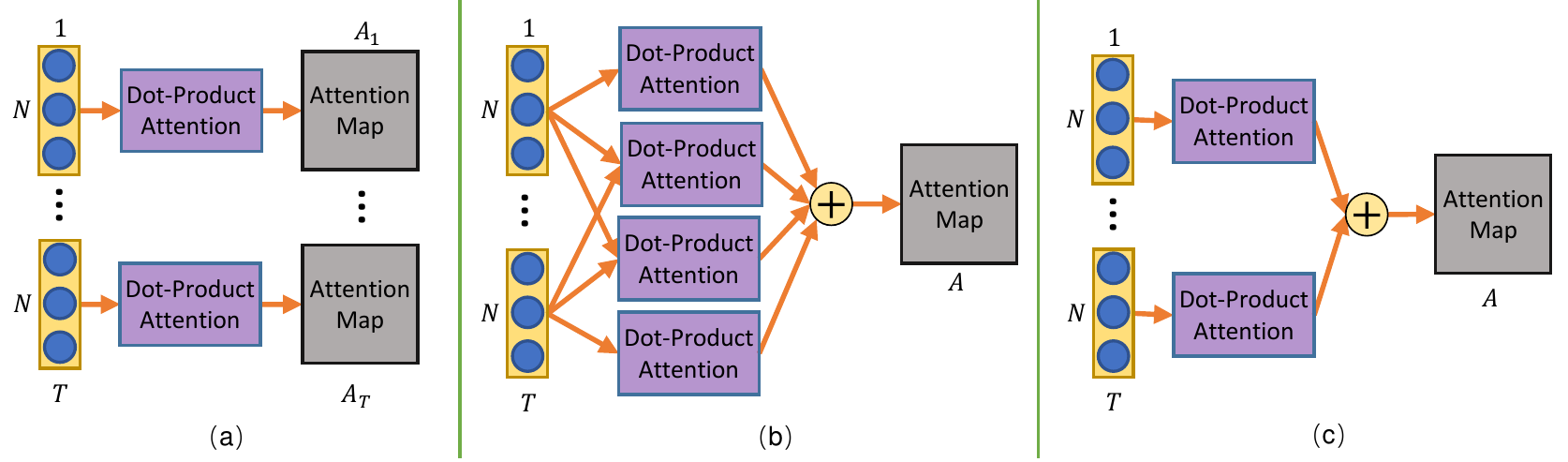}
    \caption{Illustration of the three decoupling strategies. We use the spatial attention strategy as an example and the temporal attention strategy is an analogy. N and T denote the number of joints and frames, respectively.}
    \label{fig:strategy}
\end{figure}{}

We design three strategies for decoupling as shown in Fig~\ref{fig:strategy}. 
Using the spatial attention as an example, the first strategy (Fig~\ref{fig:strategy}, a) is calculating the attention maps frame by frame, and each frame uses a unique attention map:
\begin{equation}
    A^t = softmax(\sigma(X_t)\phi(X_t)')
\end{equation}{}
where $A^t\in\mathbb{R}^{N\times N}$ is the attention map for frame $t$. $X_t\in\mathbb{R}^{N\times C}$. $\sigma$ and $\phi$ are two embedding functions. $'$ denote matrix transpose. 
This strategy only considers the dependencies of joints in a single frame thus lacks the modeling capacity. 
The computational complexity of calculating spatial attention of this strategy is $O(TN^2C)$. 
For temporal attention, the attention map of joint $n$ is $A^n\in\mathbb{R}^{T\times T}$ and the input data is $X_n\in\mathbb{R}^{T\times C}$. 
Its calculation is analogical with the spatial attention. 
Considering both the spatial and temporal attention, the computational complexity of the first strategy for all frames is $O(TN^2C+NT^2C)$.

The second strategy (Fig~\ref{fig:strategy}, b) is calculating the relations of two joints between all of the frames, which means both the intra-frame relations and the inter-frame relations of two joints are taken into account simultaneously. 
The attention map is shared over all frames. 
In formulation: 
\begin{equation}
    A^t = softmax(\sum_t^T\sum_\tau^T(\sigma(X_t)\phi(X_\tau)'))
\end{equation}{}
The computational complexity of this strategy is $O(T^2N^2C+N^2T^2C)$.

The third strategy (Fig~\ref{fig:strategy}, c) is a compromise, where only the joints in same frame are considered to calculate the attention map, but the obtained attention maps of all frames are averaged and shared.
It is equivalent to adding a time consistency restriction for attention computation, which can somewhat reduce the overfitting problem caused by the element-wise relation modeling of the second strategy.
\begin{equation}
    A^t = softmax(\sum_t^T(\sigma(X_t)\phi(X_t)'))
\end{equation}{}
By concatenating the frames into an $N\times TC$ matrix, the summation of mat-multiplications can be efficiently implemented with one big mat-multiplication operation. 
The computational complexity of this strategy is $O(TN^2C+NT^2C)$.
as shown in ablation study~\ref{sec:ablation}, we finally use the strategy (c) in the model.

\subsection{Decoupled Position encoding}
\label{sec:positionencoding}
The skeletal joints are organized as a tensor to be fed into the neural networks. 
Because there are no predefined orders or structures for each element of the tensor to show its identity (e.g., joint index or frame index), we need a position encoding module to provide unique markers for every joint. 
Following~\cite{vaswani_attention_2017}, we use the sine and cosine functions with different frequencies as the encoding functions: 
\begin{equation}
\begin{aligned}
    &PE(p, 2i) = sin(p/10000^{2i/C_{in}}) \\
    &PE(p, 2i+1) = cos(p/10000^{2i/C_{in}})    
\end{aligned}
\end{equation}
where $p$ denotes the position of element and $i$ denotes the dimension of the position encoding vector.
However, different with \cite{vaswani_attention_2017}, the input of skeletal data have two dimensions, i.e., space and time. 
One strategy for position encoding is unifying the spatial and temporal dimensions and encoding them sequentially. 
For example assuming there are three joints, for the first frame the position of joints is $1, 2, 3$, and for the second frame it is $4, 5, 6$. 
This strategy cannot well distinguish  the same joint in different frames. 
Another strategy is decoupling the process into spatial position encoding and temporal position encoding. 
Using the spatial position encoding as an example, the joints in the same frame are encoded sequentially and the same joints in different frames have the same encoding. 
In above examples, it means for the first frame the position is $1, 2, 3$, and for the second frame it is also $1, 2, 3$. 
As for the temporal position encoding, it is reversed and analogical, which means the joints in the same frame have the same encoding and the same joints in different frames are encoded sequentially. 
Finally, the position features are added to the input data as shown in Fig~\ref{fig:attention}. 
In this way, each element is aligned with an unique marker to help learning the mutual relations between the joints, and the difference between space and time is also well expressed. 

\subsection{Spatial global regularization}
\label{sec:regularization}

As explained in Sec.~\ref{sec:introduction}, each joint has a specific meaning. 
Based on this prior knowledge, we propose to add a spatial global regularization to force the model to learn more general attentions for different samples.
In detail, a global attention map ($N\times N$ matrix) is added to the attention map ($N\times N$ matrix) learned by the dot-product attention mechanism introduced in Sec.~\ref{sec:strategies}. 
The global attention map is shared for all data samples, which represents a unified intrinsic relationship pattern of the human joints. 
We set it as the parameter of the network and optimize it together with the model. 
An $\alpha$ is multiplied to balance the strength of the spatial global regularization. 
This module is simple and light-weight, but it is effective as shown in the ablation study. 
Note that the regularization is only added for spatial attention computing because the temporal dimension has no such semantic alignment property. 
Forcing a global regularization for temporal attention is not reasonable and will harm the performance. 

\subsection{Complete attention module}
\label{sec:attention}

\begin{figure}[tp]
    \centering
    \includegraphics[width=\linewidth]{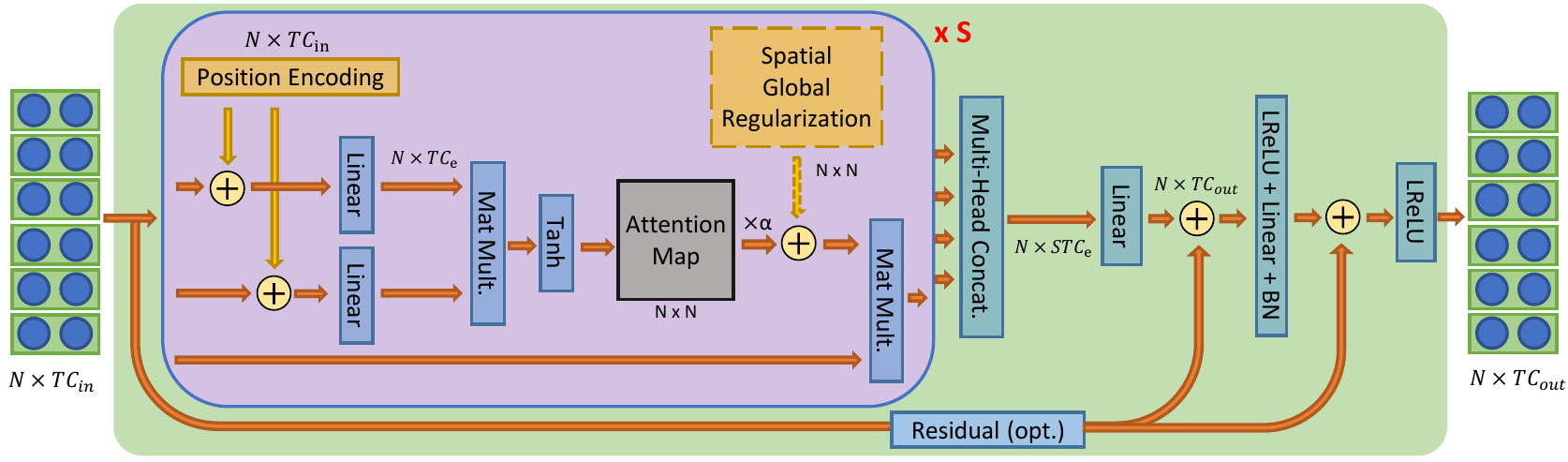}
    \caption{Illustration of the attention module. We show the spatial attention module as an example. The temporal attention module is an analogy. The purple rounded rectangle box represents a single-head self-attention module. There are totally $S$ self-attention modules, whose output are concatenated and fed into two linear layers to obtain the output. LReLU represents the leaky ReLU~\cite{maas_rectifier_2013}.}
    \label{fig:attention}
\end{figure}{}

Because the spatial attention module and the temporal attention module are analogical, we select the spatial module as an example for detailed introduction. 
The complete attention module is showed in Fig~\ref{fig:attention}. 
The procedures inside the purple rounded rectangle box illustrate the process of the single-head attention calculation. 
The input $X\in\mathbb{R}^{N\times TC_{in}}$ is first added with the spatial position encoding. 
Then it is embedded with two linear mapping functions to $X\in\mathbb{R}^{N\times TC_{e}}$. 
$C_e$ is usually small than $C_{out}$ to remove the feature redundancy and reduce the computations.
The attention map is calculated by the strategy (c) of Fig.~\ref{fig:strategy} and added with the spatial global regularization. 
Note that we found the Tanh is better than SoftMax when computing the attention map. 
We believe that it is because the output of Tanh is not restricted to positive values thus can generate negative relations and provide more flexibility. 
Finally the attention map is mat-multiplied with the original input to get the output features. 

To allow the model jointly attending to information from different representation sub-spaces, there are totally $S$ heads for attention calculations in the module. 
The results of all heads are concatenated and mapped to the output space $\mathbb{R}^{N\times TC_{out}}$ with a linear layer. 
Similar with the transformer, a point-wise feed-forward layer is added in the end to obtain the final output. 
We use the leaky ReLU as the non-linear function. 
There are two residual connections in the module as shown in the Fig~\ref{fig:attention} to stabilize the network training and integrate different features. 
Finally, all of the procedures inside the green rounded rectangle box represent one whole attention module.

\subsection{Overall architecture}
\label{sec:architecture}

\begin{figure}[tp]
    \centering
    \includegraphics[width=\linewidth]{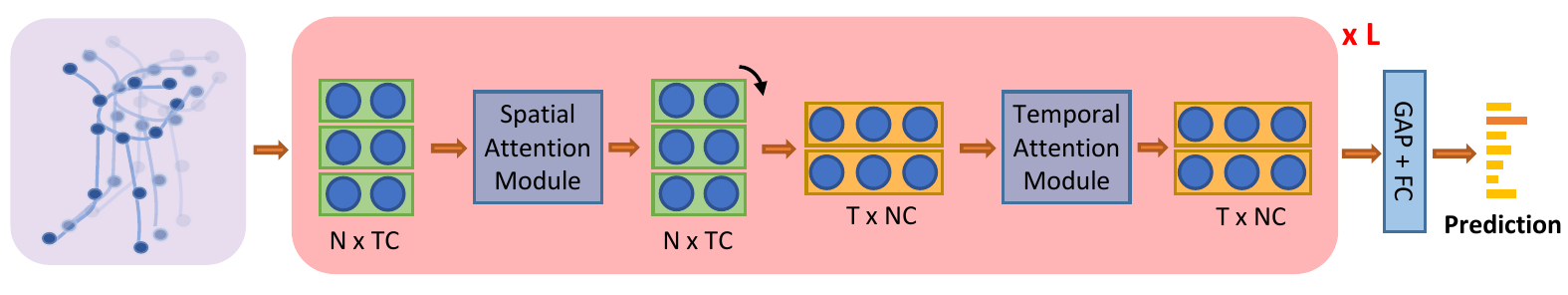}
    \caption{Illustration of the overall architecture of the DSTA-Net. 
    N, T, C denote the number of joints, frames and channels, respectively. 
    The red rounded rectangle box represents one spatial-temporal attention layer. There are totally L layers. The final output features are global-average-pooled (GAP) and fed into a fully-conected layer (FC) to make the prediction.}
    \label{fig:pipeline}
\end{figure}{}

Fig.~\ref{fig:pipeline} shows the overall architecture of our method. 
The input is a skeleton sequence with N joints, T frames and C channels. 
In each layer, we first regard the input as an $N\times TC$ matrix, i.e., N elements with $TC$ channels, and feed it into the spatial attention module (introduced in Fig.~\ref{fig:attention}) to model the spatial relations between the joints. 
Then, we transpose the output matrix and regard it as T elements each has $NC$ channels, and feed it into the temporal attention module to model the temporal relations between the frames. 
There are totally $L$ layers stacked to update features. 
The final output features are global-average-pooled and fed into a fully-connected layers to obtain the classification scores. 

\subsection{Data decoupling}
\label{sec:decouple}
\begin{figure}[tb]
    \centering
    \includegraphics[width=\linewidth]{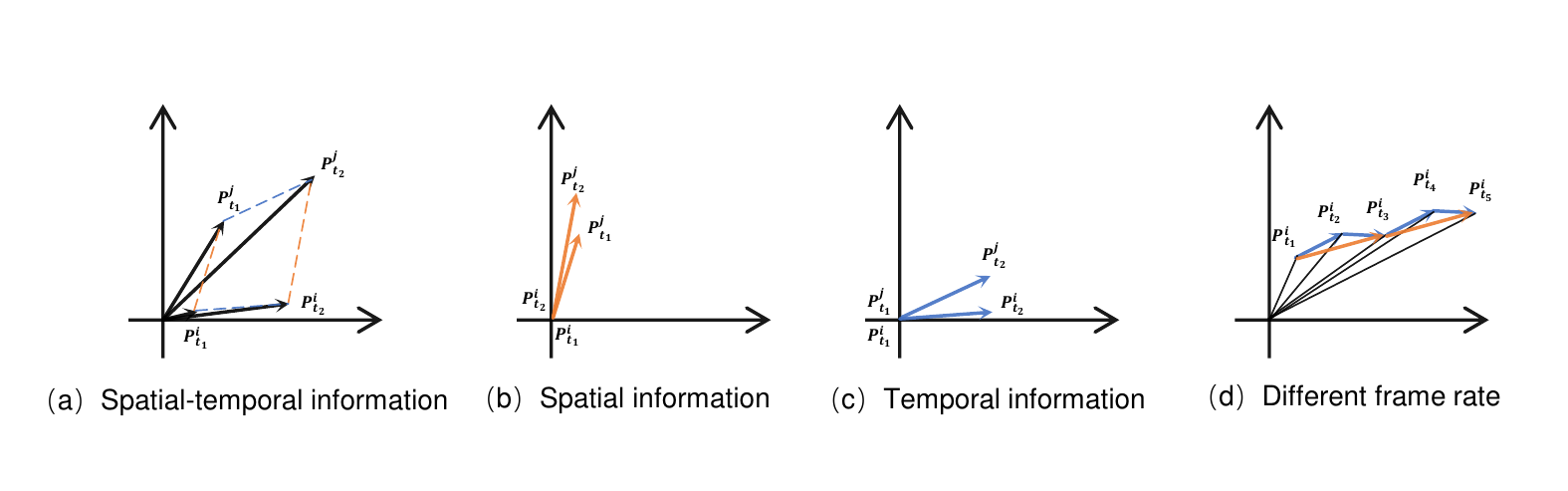}
    \caption{
    For simplicity, we draw two joins in two consecutive frames in a 2D coordinate system to illustrate the data decoupling. 
    as shown in (a), $P^{i}_{t_{k}}$ denotes the joint $i$ in frame $k$. Assume that joint $i$ and joint $j$ are the two end joints of one bone. (a) denotes the raw data, i.e., the spatial-temporal information. The orange dotted line and blue dotted line denote the decoupled spatial information and temporal information, which are showed as (b) and (c), respectively. 
    (d) illustrates the difference between the fast-temporal information (blue arrow) and the slow-temporal information (orange arrow). 
    }
    \label{fig:featurefusion}
\end{figure}{}

The action can be decoupled into two dimensions: the spatial dimension and the temporal dimension as illustrated in Fig.~\ref{fig:featurefusion} (a, b and c). 
The spatial information is the difference of two different  joints that are in the same frame, which mainly contains the relative position relationship between different joints. 
To reduce the redundant information, we only calculate the spatial information along the human bones. 
The temporal information is the difference of the two joints with same spatial meaning in different frames, which mainly describes the motion trajectory of one joint along the temporal dimension. 
When we recognize the gestures like ``Point with one finger'' versus ``Point with two finger'', the spatial information is more important. 
However, when we recognize the gestures like ``waving up'' versus ``waving down'', the temporal information will be more essential. 


Besides, for temporal stream, different actions have different sensibilities of the motion scale. 
For some actions such as ``clapping" versus ``put two hands together", the short-term motion detail is essential. But for actions like ``waving up" versus ``waving down", the long-term movement trend is more important. 
Inspired by~\cite{feichtenhofer_slowfast_2019}, we propose to calculate the temporal motion with both the high frame-rate sampling and the low frame-rate sampling as shown in Fig.~\ref{fig:featurefusion} (d). 
The generated two streams are called as the fast-temporal stream and the slow-temporal stream, respectively. 

Finally, we have four streams all together, namely, spatial-temporal stream (original data), spatial stream, fast-temporal stream and slow-temporal stream. 
We separately train four models with the same architecture for each of the streams. 
The classification scores are averaged to obtain the final result.

\section{Experiments}
To verify the generalization of the model, we use two datasets for hand gesture recognition (DHG~\cite{de_smedt_skeleton-based_2016} and SHREC~\cite{de_smedt_shrec17_2017}) and two datasets for human action recognition (NTU-60~\cite{shahroudy_ntu_2016} and NTU-120~\cite{liu_ntu_2019}). We first perform exhaustive ablation studies on SHREC to verify the effectiveness of the proposed model components. Then, we evaluate our model on all four datasets to compare with the state-of-the-art methods.

\subsection{Datasets}
\textbf{DHG:} DHG~\cite{de_smedt_skeleton-based_2016} dataset contains 2800 video sequences of 14 hand gestures performed 5 times by 20 subjects. They are performed in two ways: using one finger and the whole hand. So it has two benchmarks: 14-gestures for coarse classification and 28-gestures for fine-grained classification. The 3D coordinates of 22 hand joints in real-world space is captured by the Intel Real-sense camera. It uses the leave-one-subject-out cross-validation strategy for evaluation. 

\textbf{SHREC:} SHREC~\cite{de_smedt_shrec17_2017} dataset contains 2800 gesture sequences performed 1 and 10 times by 28 participants in two ways like the DHG dataset. It splits the sequences into 1960 train sequences and 840 test sequences. The length of sample gestures ranges from 20 to 50 frames. This dataset is used for the competition of SHREC'17 in conjunction with the Euro-graphics 3DOR'2017 Workshop. 

\textbf{NTU-60:} NTU-60~\cite{shahroudy_ntu_2016} is a most widely used in-door-captured action recognition dataset, which contains 56,000 action clips in 60 action classes. The clips are performed by 40 volunteers and is captured by 3 KinectV2 cameras with different views. This dataset provides 25 joints for each subject in the skeleton sequences. It recommends two benchmarks: cross-subject (CS) and cross-view (CV), where the subjects and cameras used in the training/test splits are different, respectively. 

\textbf{NTU-120:} NTU-120~\cite{shahroudy_ntu_2016} is similar with NTU-60 but is larger. It contains 114,480 action clips in 120 action classes. The clips are performed by 106 volunteers in 32 camera setups. It recommends two benchmarks: cross-subject (CS) and cross-setup (CE), where cross-setup means using samples with odd setup IDs for training and others for testing. 

\subsection{Training details}
To show the generalization of our methods, we use the same configuration for all experiments. 
The network is stacked using 8 DSTA blocks with 3 heads. The output channels are 64, 64, 128, 128, 256, 256, 256 and 256, respectively. 
The input video is randomly/uniformly sampled to 150 frames and then randomly/centrally cropped to 128 frames for training/test splits. 
For fast-temporal features, the sampling interval is 2. 
When training, the initial learning rate is 0.1 and is divided by 10 in 60 and 90 epochs. 
The training is ended in 120 epochs. 
Batch size is 32. Weight decay is $0.0005$. 
We use the stochastic gradient descent (SGD) with Nesterov momentum ($0.9$) as the optimizer and the cross-entropy as the loss function.

\subsection{Ablation studies}
\label{sec:ablation}
In this section, we investigate the effectiveness of the proposed components of the network and different data modalities. 
We conduct experiments on SHREC dataset. 
Except for the explored object, other details are set the same for fair comparison. 

\subsubsection{Network architectures}
We first investigate the effect of the position embedding. 
as shown in Tab.~\ref{tab:ablation_model}, removing the position encoding will seriously harm the performance. Decoupling the spatial and temporal dimension (DPE) is better than not (UPE). 
This is because the spatial and temporal dimensions actually have different properties and treat them equivalently will confuse the model.

Then we investigate the effect of the proposed spatial global regularization (SGR). 
By adding the SGR, the performance is improved from 94.3\% to 96.3\%, but if we meanwhile regularize the temporal dimension, the performance drops. 
This is reasonable since there are no specific meanings for temporal dimension and forced learning of a unified pattern will cause the gap between the training set an testing set. 

Finally, we compare the three strategies introduced in Fig.~\ref{fig:strategy}. 
It shows that the strategy (a) obtains the lowest performance. 
We conjecture that it dues to the fact that it only considers the intra-frame relations and ignore the inter-frame relations. 
Modeling the inter-frame relations exhaustively (strategy b) will improve the performance and a compromise (c) obtains the best performance. 
It may because that the compromise strategy can somewhat reduce the overfitting problem. 
\begin{table}[!htp]
    \centering
    \caption{Ablation studies for architectures of the model on the SHREC dataset. 
    ST-ATT-c denotes the \textbf{s}patial \textbf{t}emporal \textbf{att}ention networks with attention type \textbf{c} introduced in Fig~\ref{fig:strategy}. 
    PE denotes \textbf{p}osition \textbf{e}ncoding. 
    UPE/DPE denote using \textbf{u}nified/\textbf{d}ecoupled encoding for spatial and temporal dimensions. 
    STGR denotes \textbf{s}patial-\textbf{t}emporal \textbf{g}lobal \textbf{r}egularizations for computing attention maps. }
    \renewcommand\tabcolsep{5.0pt} 
    \begin{tabular}{lc}
    \hline
    Method & Accuracy \\
    \hline
    ST-Att-c w/o PE       & 89.4 \\
    ST-Att-c + UPE        & 93.2 \\
    ST-Att-c + DPE        & 94.5 \\
    \hline
    ST-Att-c + DPE + SGR   & \textbf{96.3} \\
    ST-Att-c + DPE + STGR  & 94.6 \\
    \hline
    ST-Att-a  + DPE + SGR    & 94.6 \\
    ST-Att-b  + DPE + SGR    & 95.1 \\
    \hline
    \end{tabular}
    \label{tab:ablation_model}
\end{table}{}

We show the learned attention maps of different layers (layer \#1 and layer \#8) in Fig.~\ref{fig:attentionmap}. 
Other layers are showed in supplement materials. 
It shows that the attention maps learned in different layers are not the same because the information contained in different layers has distinct semantics. 
Besides, it seems the model focuses more on the relations between the tips of the fingers (T4, I4, M4, R4) and wrist, especially in the lower layers. 
This is intuitive since these joints are more discriminative for human to recognize gestures. 
On the higher layers, the information are highly aggregated and the difference between each of the joints becomes unapparent, thus the phenomenon also becomes unapparent.

\begin{figure}[!htb]
    \centering
    \includegraphics[width=\linewidth]{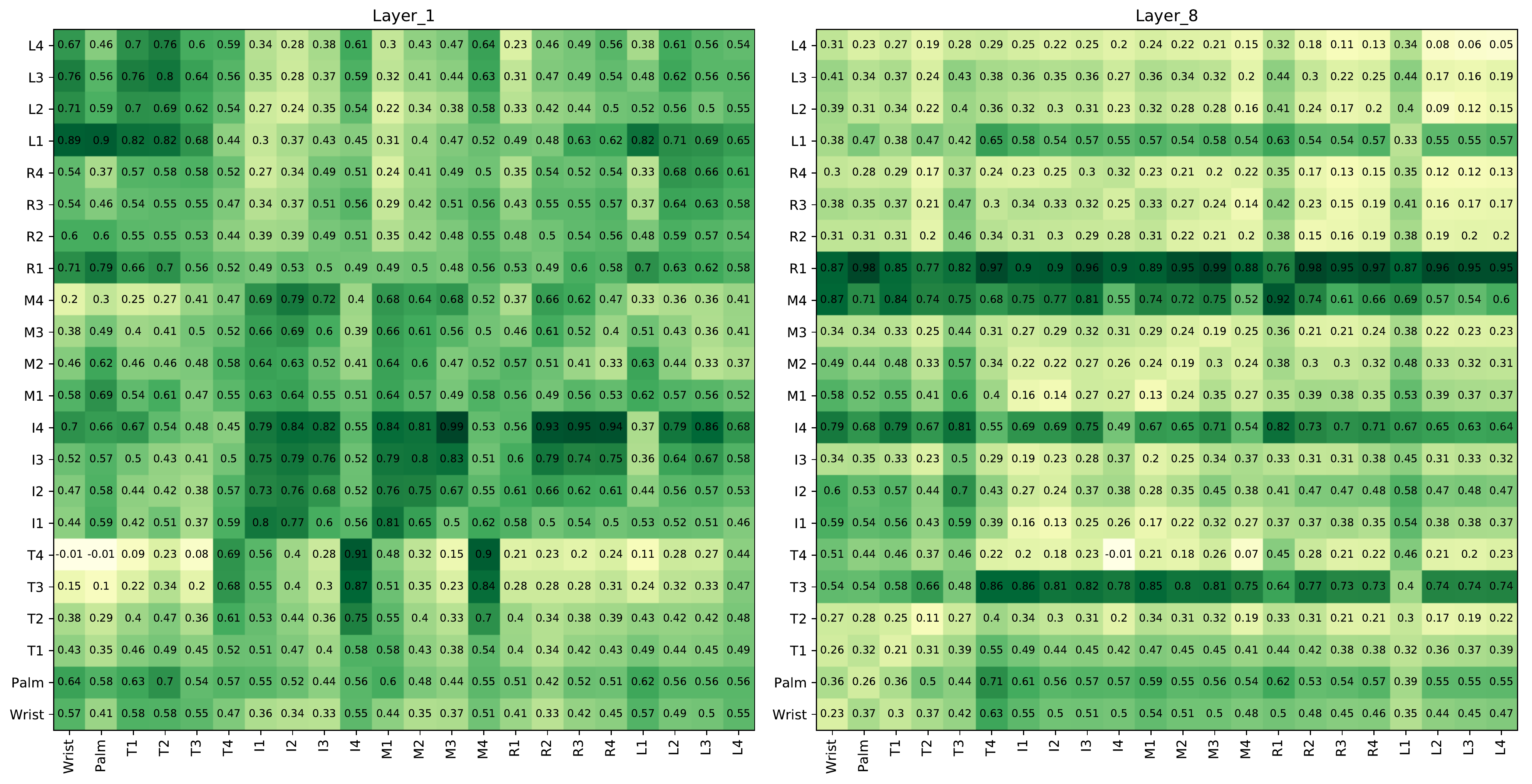}
    \caption{Examples of the learned attention maps for different layers. T, I, M, R and L denote thumb, index finger, middle finger, ring finger and little finger, respectively. As for articulation, T1 denotes the base of the thumb and T4 denote the tip of the thumb.}
    \label{fig:attentionmap}
\end{figure}{}

\subsubsection{Data decoupling}
To show the necessity of decoupling the raw data into four streams as introduced in Sec.~\ref{sec:decouple}, we show the results of using four streams separately and the result of fusion in Tab.~\ref{tab:ablation_fusion}.
It shows that the accuracies of decoupled streams are not as good as the raw data because some of the information is lost. 
However, since the four streams focus on different aspects and are complementary with each other, when fusing them together, the performance is improved significantly. 

\begin{table}[!htp]
    \centering
    \caption{Ablation studies for feature fusion on the SHREC dataset. Spatial-temporal denotes the raw data, i.e., the joint coordinates. Other types of features are introduced in Sec.~\ref{sec:decouple}.}
    \renewcommand\tabcolsep{5.0pt} 
    \begin{tabular}{lc}
    \hline
    Method & Accuracy \\
    \hline
    spatial-temporal        &  96.3 \\
    spatial         &  95.1 \\
    fast-temporal   &  94.5\\
    slow-temporal   &  93.7 \\
    \hline
    Fusion          &  \textbf{97.0}\\
    \hline
    \end{tabular}
    \label{tab:ablation_fusion}
\end{table}{}

\begin{figure}[!htb]
    \centering
    \includegraphics[width=\linewidth]{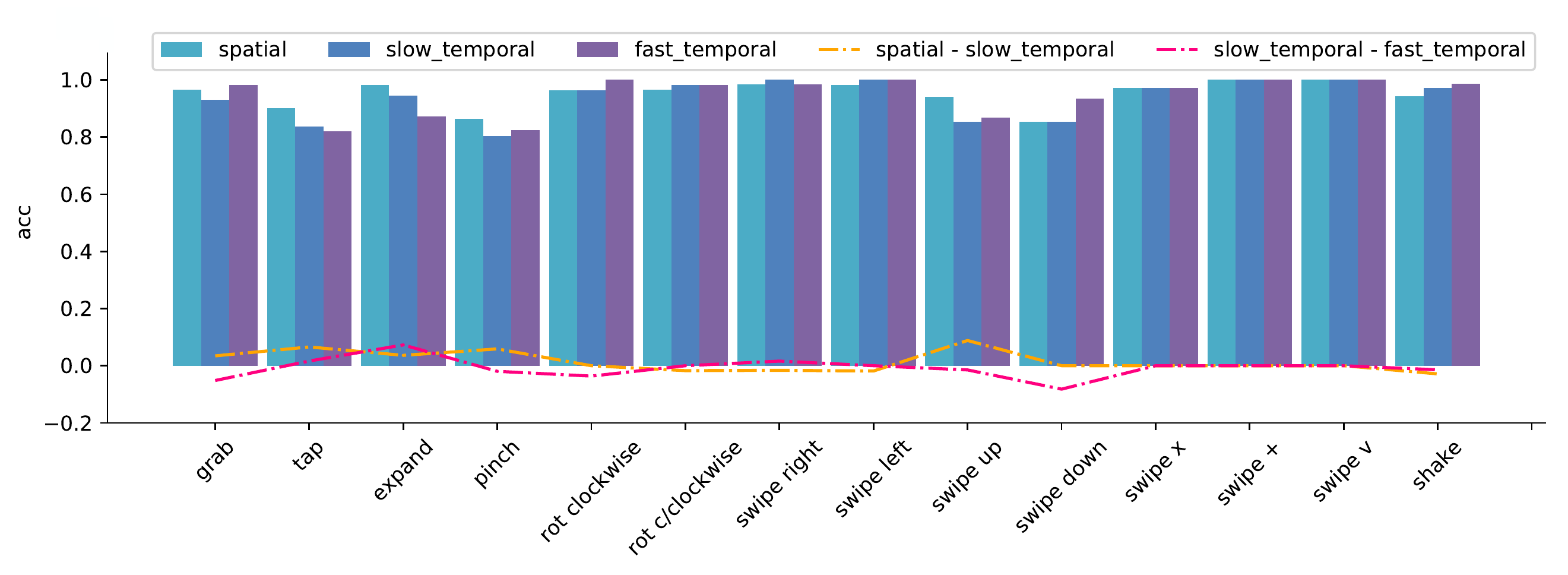}
    \caption{Per-class accuracies for different modalities on SHREC-14 dataset. The dotted lines shows the difference between two modalities.}
    \label{fig:featurefusionacc}
\end{figure}{}

As shown in Fig.~\ref{fig:featurefusionacc}, We plot the per-class accuracies of the four streams to show the complementarity clearly. 
We also plot the difference of accuracies between different streams, which are represented as the dotted lines.  
For spatial information versus temporal information, it (orange dotted lines) shows that the network with spatial information obtains higher accuracies mainly in classes that are closely related with the shape changes such as ``grab'', ``expand'' and ``pinch'', and the network with temporal information obtains higher accuracies mainly in classes that are closely related with the positional changes such as ``swipe'', ``rot'' and ``shake''. 
As for different frame-rate sampling, it (red dotted lines) shows that the slow-temporal performs better for classes of ``expand'', ``tap'', etc, and the fast-temporal performs better for classes of ``swipe'', ``rot'', etc. 
These phenomenons verify the complementarity of the four modalities.

\begin{table}[!htp]
    \centering
    \caption{Recognition accuracy comparison of our method
and state-of-the-art methods on SHREC dataset and DHG dataset.}
    \renewcommand\tabcolsep{5.0pt} 
    \begin{tabular}{l|c|c|c|c|c}
    \hline
    \multirow{2}*{Method} & \multirow{2}*{Year} & \multicolumn{2}{c|}{SHREC} & \multicolumn{2}{c}{DHG}\\ 
    \cline{3-4} \cline{5-6} 
    ~ & ~ & 14 gestures & 28 gestures & 14 gestures & 28 gestures\\
    \hline
    ST-GCN~\cite{yan_spatial_2018} & 2018 &  92.7 & 87.7 &  91.2 & 87.1 \\
    STA-Res-TCN~\cite{hou_spatial-temporal_2018} & 2018 & 93.6 & 90.7 & 89.2 & 85.0 \\
    \hline
    ST-TS-HGR-NET~\cite{nguyen_neural_2019} & 2019 &  94.3 & 89.4  &  87.3 & 83.4\\
    DG-STA.~\cite{chen_construct_2019} & 2019 & 94.4 & 90.7 & 91.9 & 88.0 \\
    \hline
    \textbf{DSTA-Net(ours)} & - & \textbf{97.0} & \textbf{93.9}  & \textbf{93.8} & \textbf{90.9}\\
    \hline
    \end{tabular}
    \label{tab:shrec}
\end{table}{}


\begin{table}[htb]
  \centering
  \caption{Recognition accuracy comparison of our method
and state-of-the-art methods on NTU-60 dataset. CS and CV denote the cross-subject and cross-view benchmarks, respectively. }
    \renewcommand\tabcolsep{5.0pt} 
	\begin{tabular}{l|c|c|c}
		\hline
		Methods  & Year   &CS (\%)& CV (\%)    \\
        \hline
        ST-GCN~\cite{yan_spatial_2018} & 2018 & 81.5  &      88.3  \\
        SRN+TSL~\cite{si_skeleton-based_2018} & 2018 & 84.8 & 92.4 \\
        \hline
        2s-AGCN~\cite{shi_two-stream_2019} & 2019  & 88.5  & 95.1\\
        DGNN~\cite{shi_skeleton-based_2019}  & 2019 & 89.9  & 96.1\\
        \hline
        NAS~\cite{peng_learning_2020}  & 2020 & 89.4  & 95.7\\
		\hline
        \textbf{DSTA-Net(ours)} & - & \textbf{91.5} & \textbf{96.4} \\
		\hline
	\end{tabular}
  \label{tab:ntu-60}
\end{table}

\begin{table}[htb]
  \centering
    \caption{Recognition accuracy comparison of our method
and state-of-the-art methods on NTU-120 dataset. CS and CE denote the cross-subject and cross-setup benchmarks, respectively. }
    \renewcommand\tabcolsep{5.0pt} 
	\begin{tabular}{l|c|c|c}
		\hline
		Methods  & Year   &CS (\%)& CE (\%)    \\
        \hline
        Two-Stream Attention LSTM~\cite{liu_skeleton-based_2017} & 2017 & 61.2 &    63.3  \\
        Body Pose Evolution Map~\cite{liu_recognizing_2018}  & 2018 & 64.6 &    66.9  \\
        SkeletonMotion~\cite{caetano_skelemotion:_2019}  & 2019 & 67.7 &    66.9  \\
        \hline
        \textbf{DSTA-Net(ours)} & - & \textbf{86.6} & \textbf{89.0} \\
		\hline
	\end{tabular}
  \label{tab:ntu-120}
\end{table}

\subsection{Comparison with previous methods}
We evaluate our model with state-of-the-art methods for skeleton-based action recognition on all four datasets, where our model significantly outperforms the other methods. 
Due to the space restriction, we only show some representative works, where more comparisons are showed in supplement materials. 
On SHREC/DHG datasets for skeleton-based hand gestures recognition (Tab.~\ref{tab:shrec}), our model brings $2.6\%$/$1.9\%$ and $3.2\%$/$2.9\%$ improvements for 14-gestures and 28-gestures benchmarks compared with the state-of-the-arts.
Note that the state-of-the-art accuracies are already very high ($94.4\%$/$91.9\%$ and $90.7\%$/$88.0\%$ for 14-gestures and 28-gestures, respectively), but our model still obtains remarkable performance. 
On NTU-60 dataset (Tab.~\ref{tab:ntu-60}), our model obtains $1.6\%$ and $0.3\%$ improvements. 
The performance of CV benchmark is nearly saturated. For both CS and CV benchmarks, we visualize the wrong examples and find that it is even impossible for human to recognize many examples using only the skeletal data. For example, for the two classes of reading and writing, the humans are both in a same posture (standing or sitting) and holding a book. The only difference is whether there is a pen in the hand, which cannot be captured through the skeletal data. 
On NTU-120 dataset (Tab.~\ref{tab:ntu-120}), our model also achieves state-of-the-art performance. Since this dataset is released recently, our method can provide a new baseline on it.

\section{Conclusion}
In this paper, we propose a novel decoupled spatial-temporal attention network (DSTA-Net) for skeleton-based action recognition. 
It is a unified framework based solely on attention mechanism, with no needs of designing hand-crafted traversal rules or graph topologies. 
We propose three techniques in building DSTA-Net to meet the specific requirements for skeletal data, including spatial-temporal attention decoupling, decoupled position encoding and spatial global regularization. 
Besides, we introduce a skeleton-decoupling method to emphasize the spatial/temporal variations and motion scales of the skeletal data, resulting in a more comprehensive understanding for human actions and gestures. 
To verify the effectiveness and generalizability of the DSTA-Net, extensive experiments are conducted on four large datasets for both gesture and action recognition, where the DSTA-Net achieves the state-of-the-art performance on all of them. 

\bibliographystyle{splncs04}
\bibliography{references}

\end{document}